\documentclass[10pt, conference, compsocconf]{IEEEtran}
\ifCLASSINFOpdf
\else
\fi
\usepackage{graphicx}

\begin{document}
\title{Growing Deep Forests Efficiently with Soft Routing and Learned Connectivity}


\author{
\IEEEauthorblockN{Jianghao Shen, Sicheng Wang, and Zhangyang Wang}
\IEEEauthorblockA{Department of Computer Science and Engineering\\
Texas A\&M University, College Station, TX, 77840\\
Email: \{nie, sharonwang, atlaswang\}@tamu.edu}\\   %<------ Line breaks in the current column
% \IEEEauthorblockN{Zhangyang Wang}
% \IEEEauthorblockA{Department of Computer Science\\
% and Engineering\\
% Texas A\&M University\\
% College Station, TX, 77840\\
% Email:atlaswang@tamu.edu}
% \and
% \IEEEauthorblockN{Sicheng Wang}
% \IEEEauthorblockA{Department of Computer Science\\
% and Engineering\\
% Texas A\&M University\\
% College Station, TX, 77840\\
% Email: sharonwang@tamu.edu}\\  %<------- Extra vertical space

                %<-----------

}

%\vspace{0.05in}


%Telephone: (800) 555--1212, Fax: (888) 555--1212}


\maketitle

\begin{abstract}
Despite the latest prevailing success of deep neural networks (DNNs), several concerns have been raised against their usage, including the lack of intepretability the gap between DNNs and other well-established machine learning models, and the growingly expensive computational costs. A number of recent works \cite{zhou2017deep,hettinger2017forward,miller2017forward} explored the alternative to sequentially stacking decision tree/random forest building blocks in a purely feed-forward way, with no need of back propagation. Since decision trees enjoy inherent reasoning transparency, such deep forest models can also facilitate the understanding of the internal decision making process. This paper further extends the deep forest idea in several important aspects. Firstly, we employ a probabilistic tree whose nodes make probabilistic routing decisions, a.k.a., “soft routing”, rather than hard binary decisions. Besides enhancing the flexibility, it also enables non-greedy optimization for each tree. Second, we propose an innovative topology learning strategy: every node in the ree now maintains a new learnable hyperparameter indicating the probability that it will be a leaf node. In that way, the tree will jointly optimize both its parameters and the tree topology during training. Experiments on the MNIST dataset demonstrate that our empowered deep forests can achieve better or comparable performance than \cite{zhou2017deep,miller2017forward} , with dramatically reduced model complexity. For example, our model with only 1 layer of 15 trees can perform  comparably with the model in \cite{miller2017forward} with 2 layers of 2000 trees each.

\end{abstract}

\begin{IEEEkeywords}
soft routing; topology learning; deep forest;

\end{IEEEkeywords}

% \fi
\IEEEpeerreviewmaketitle

\section{Introduction}
Although deep neural networks (DNNs) haven been successful in many computer vision and pattern recognition tasks, several limitations have prevented their wider usage. For example, due to the high computational cost for training and inference, it is difficult to deploy large DNN models to devices that have limited computing resources \cite{wu2018deep}. The lack of interpretability makes DNN unsuitable for real world tasks where the decision-making explainability is crucial. Since DNNs essentially stack repeatedly basic computation units - a perceptron, it is natural to ask whether replacing this atomic unit with other modules will bring any positive surprise. 

The decision tree is known for its model transparency and efficient training method. Can we build a deep model that uses decision tree as the basic computation unit? That idea has been explored in several recent works and witnesses preliminary success. \cite{zhou2017deep} proposed a deep random forest cascade, where each layer of the network is an ensemble
of a set of random forests, and the concatenation of the
current forests' outputs becomes the input of the next layer. \cite{miller2017forward} improved over \cite{zhou2017deep} by using a set of decision trees per layer, but concatenating the outputs of the current layer with the original input to form a new representation for the next layer. Both \cite{zhou2017deep} and \cite{miller2017forward} have reported comparable results on the MNIST dataset with respect to the DNN models, accompanied with
significant reduction of training time and lighter-weight parameters. However, their models still demand thousands of trees at each layer, because the expressive power of a single decision tree is limited.

To address these problems, we propose to advance the existing deep forest model in several aspects. First, in order to
facilitate end-to-end optimization,
we replace the hard splitting criterion in a classical decision tree with a \textit{soft routing} mechanism, where
the routing criteria at each internal node of the decision
tree is a sigmoid function. The error signal can thus be
propagated through the tree by gradient descent.  Second, we allow the tree to learn its topology globally too, by adding a \textit{leafness parameter} for each tree node. Therefore every node can simultaneously learn whether to stay as a leaf or an inner node \cite{irsoy2014budding}.

\section{Related Work}

In this section, we introduce the related concepts of soft tree and budding tree, which enable soft routing and optimal tree structure learning.

\subsection{Soft Decision Tree}

For the traditional decision tree model, each inner tree node makes hard decision, sending the input to one of its child node, where the soft decision tree \cite{irsoy2012soft} makes a soft decision, i.e. send the input to both of its child nodes, with probability calculated by a gating function. Specifically, for an inner node m, the output of m is:

% As opposed to the hard decision node which redirects
% instances to one of its children depending on its splitting
% criteria, a soft decision node \cite{irsoy2012soft} redirects instance to all its
% children with probabilities calculated by a gating function
% . For the scenario of a binary non-leaf node m where we have
% left and right children, the output of node m is:

\begin{equation}
\begin{array}{l}\label{privacy}
y_m(x)=g_m(x)y_{ml}(x)+(1-g_m(x))y_{mr}(x)\\
\end{array}
\label{equation_one}
\end{equation}

Where x is the input, $y_{ml}$ and $y_{mr}$ represent the output of m's left child node and right child node, respectively. $g_m$ represents the gating function, its sigmoidal functional form is:

\begin{equation}
\begin{array}{l}\label{privacy_one}
g_m(x) = \frac {1} {1 + exp^{-[w_m^Tx + w_{m0}]}}\\
\end{array}
\end{equation}

$w_m$ is the weight vector of the gating function, $w_{m0}$ is the bias. Equation \ref{equation_one} is calculated recursively, until the leaf nodes are reached.
When m is a leaf node, its output is:

\begin{equation}
\begin{array}{l}\label{privacy}
y_m(x) = \rho_m
\end{array}
\end{equation}

Where $\rho_m$ is a learnable vector, its length equals the class number of the classification task. 

During training, initially the soft tree model starts with one node, to decide whether the tree will grow, two candidate child nodes will be added, then after training for several epochs, if there is performance gain compared to the single node, then the tree will grow, and following a similar process, the two child nodes will keep growing if needed, the tree stops growing if it has reached a pre-specified depth or the performance gain is below some threshold.
% During training, the  starts with one node uses its $\rho_m$ to learn a constant classifier. Then the node will splits as long as there is performance improvement. The parameters of the gating function and the $\rho_m$ vector can be optimized through gradient descent method, and the loss function for the model is cross entropy for classification tasks, and mean square loss for regression tasks.

Since the soft tree is growing in a greedy fashion in that it only optimizes the parameters of the node that will potentially grow, it is still not able to learn a global optimal topology.

\subsection{Budding Tree}
To address the limitation of soft tree, a modified version of the model - budding tree \cite{irsoy2014budding}, is introduced. The contribution of budding tree is that it treats the leaf/non-leaf state of all the tree nodes as a learnable leafness parameter $\gamma$: 

\begin{equation}
\begin{array}{l}\label{privacy}
y_m(x)=(1-\gamma_m)[g_m(x)y_{ml}(x)+(1-g_m(x))y_{mr}(x)] \\
\qquad \qquad +\gamma_m\rho_m\\
\end{array}
\end{equation} 

$\gamma$ indicates the probability that a tree node is a leaf node, thus, the final output of node m will be the combination of its role as a leaf node and non-leaf node, weighted by its leafness probability $\gamma$ and non leafness probability $(1 - \gamma)$. The Recursion ends when a node having leafness parameter 1 is encountered. Since now every node in the tree can learn to stay as a leaf or non-leaf through gradient descent, the budding tree is able to globally optimize its topology.

\subsection{Distributed Decision Trees}
One of the advantages of soft routing is that both the left branch and right branch contribute to the final classification vector, leading to a richer representation compared to the single hard selection. In order to go one step further to strengthen this merit, the distributed decision tree \cite{irsoy2014distributed} is developed. Unlike the budding tree, where there is a probabilistic constraint the the weights of left child and right child sum to one, the distributed decision trees set parameters for each child independently:

\begin{equation}
\begin{array}{l}\label{privacy}
y_m(x)=(1-\gamma_m)[g_m(x)y_{ml}(x)+h_m(x)y_{mr}(x)]+\gamma_m\rho_m\\
\end{array}
\end{equation} 

where $g_m$ and $h_m$ are calculated by two independent gating functions with different weight vectors, follow equation \ref{privacy_one}. The motivation behind this change is that since now the left and right child calculating their activations independently, they are not correlated with each other, and thus giving the tree more degrees of freedom in expressing rich representations.

\section{Our Approach}

In this section we introduce two extensions we made to the budding tree model, the first one is adding multiple filters at innder tree node, and the second one is to develop a deep fores-like model by using budding tree as a base building block.

\subsection{Multiple weight maps at each node}\label{AA}

\begin{figure*}[ht]
	\begin{center}
		\includegraphics[width=0.7\linewidth]{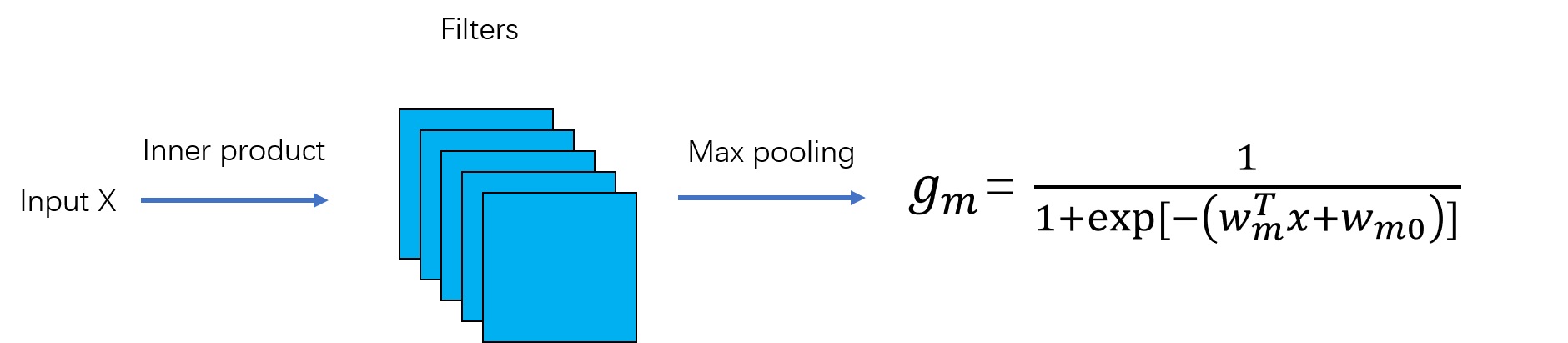}
	\end{center}
	\caption{The flow chart of multi-filter}
	\label{fig:fig1}
\end{figure*}

For the budding tree model, each inner node decide the probability of sending the input by calculating the similarity between the input and its weight map. Thus, the weight map is designed to learn distinguishable feature that can help to decide which of the two child nodes the input is more likely to go. However, it is difficult for a inner node to summarize all the distinguishable contents of the images in one weight map, thus, we propose to add multiple weight maps for each inner node, so that it can select the feature that is most similar to the current input, and decide the routing probability. The flow chart of this operation is shown at Figure \ref{fig:fig1}.
% Typical convolutional neural networks have myriad of filters per layer, we are inspired in similar way. In the budding tree, the weight of each node serves as
%a filter, to determine the likelihood of sending the current
%input to its left or right children. The learned filter indicates
%what feature the node uses for redirecting the input. However, since each node only has one filter, it is sometimes difficult to learn distinguishable feature in one filter, thus, we add multiple filters for each node, during training, the filter that achieves the largest dot product value with the input will be activated, and the dot product value will be the corresponding weight. The flow chat of the operation is shown at Figure \ref{fig:fig1}.

\begin{figure*}[ht]
	\begin{center}
		\includegraphics[width=0.7\linewidth]{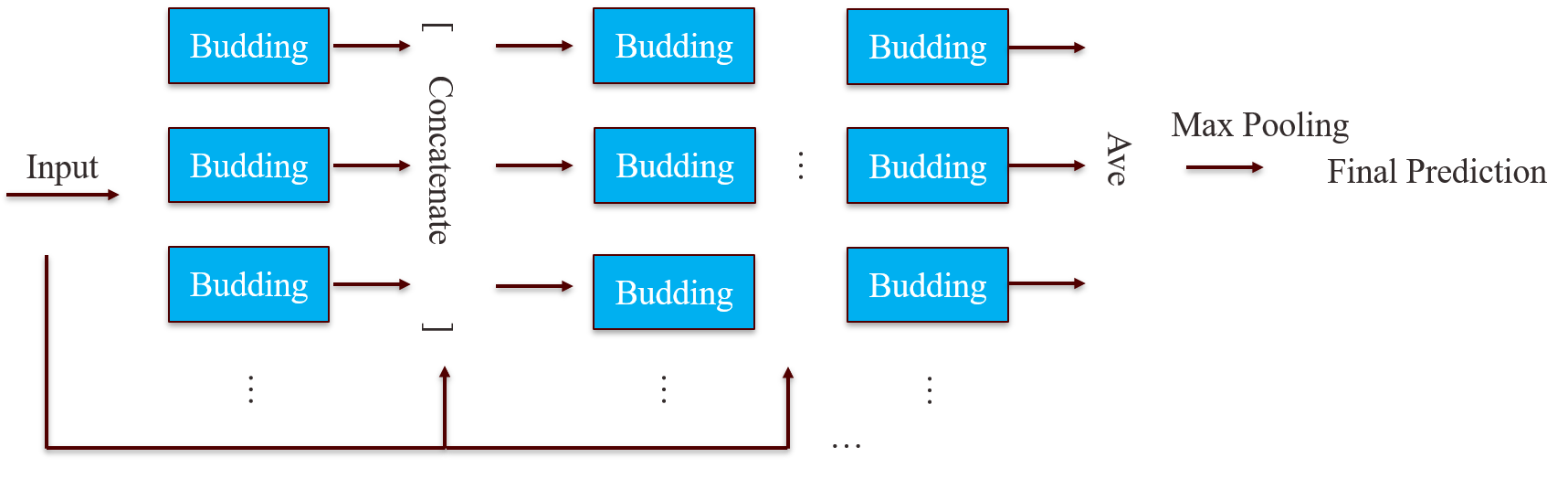}
	\end{center}
	\caption{The architecture of deep budding forest}
	\label{fig:fig2}
\end{figure*}

\subsection{Deep Budding Forest}\label{AA}

Since our goal is to build a novel deep architecture based on the modified soft decision tree, we propose to construct a deep budding forest model, the structure of the model is shown at Figure \ref{fig:fig2}. Specifically, each layer of the model consists of a series of budding trees, the input is sent to each of them for inference, the inference results for that layer will be concatenated together with the original input. The motivation that we concatenate the original input with the intermediate inference results is to prevent information loss, similar to \cite{huang2017densely} and \cite{zhou2017deep}. When the final layer is reached, all the inference results will be averaged, then the class with the maximum probability will be the output.

\section{Experimental Results}
\begin{table*}[ht]
	\caption{Experimental results and comparison with the deep forest model of forward thinking paper, and Deep Conv Net} % title of Table
	\centering % used for centering table
	\begin{tabular}{|c|c|c|c|c|} % centered columns (4 columns)
		\hline %inserts double horizontal lines
		{} & Deep Distributed Forest & Deep Budding Forest & Feed Forward Deep Forest \cite{miller2017forward} & Deep Conv Net \cite{ciresan2011flexible} \\ [0.5ex] % inserts table
		%heading
		\hline % inserts single horizontal line
		Accuracy & 98.6\% & 98.3\% & 97.58\% & 99.65\% \\ % inserting body of the table
		\hline % inserts single horizontal line
		Layers & 3 & 3 & 2 & 7 \\
		\hline % inserts single horizontal line
		Settings & 15 trees/layer & 15 trees/layer & 2000 hard trees/layer & N/A \\
		\hline % inserts single horizontal line
        Parameters & around 3175200 &  around 1587600 & around 4096000 & N/A
        \\
        \hline
		Data augmentation  & All & All & Shift & All \\
		\hline % inserts single horizontal line
	\end{tabular}
	\label{table:comp} % is used to refer this table in the text
\end{table*}

\begin{figure*}[ht]
	\begin{center}
		\includegraphics[width=0.7\linewidth]{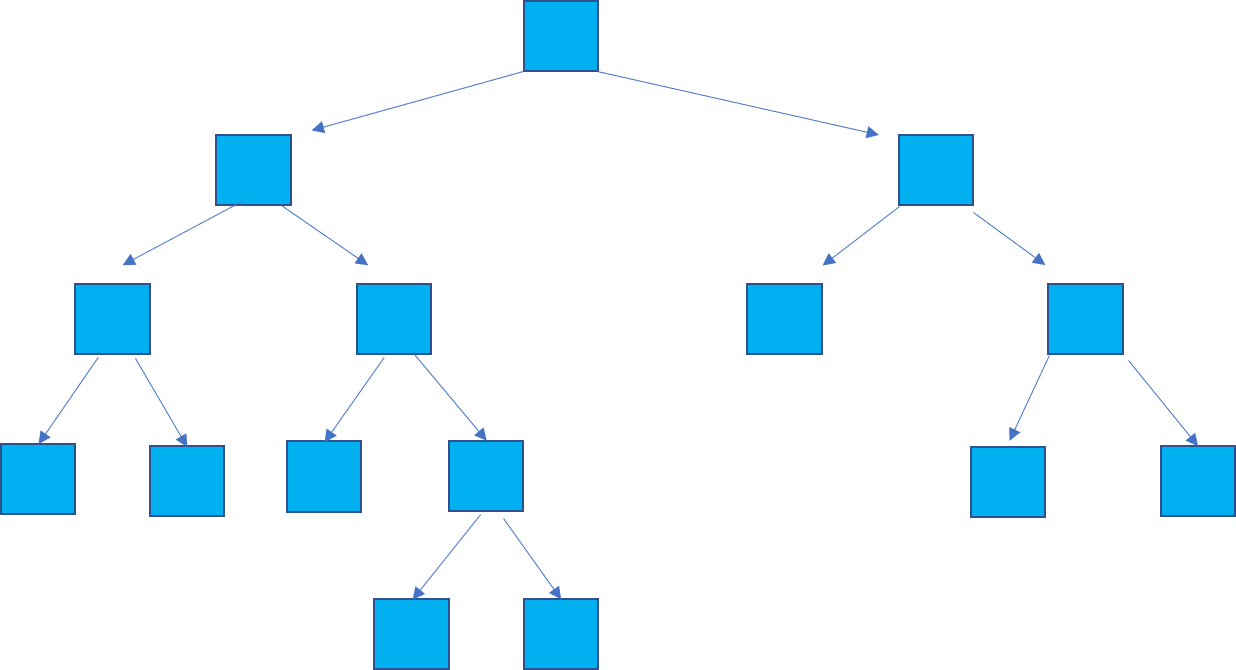}
	\end{center}
	\caption{The learned structure of a budding tree with depth 5 at one layer. Each blue box represents a tree node, it can be observed that most of the tree nodes stop growing at depth 4, indicating that the budding tree is able to learn a compact model that optimally solve the problem at hand. }
	\label{fig:fig3}
\end{figure*}

We evaluate the performance of our deep budding forest on the MNIST dataset. Our model architecture is a three layer budding forest, each layer has 15 budding trees. The model is trained layer by layer, and for each layer, the budding trees are trained independently. Thus, we construct 15 unique training set from the original MNIST training set by sampling with replacement. We also add data augmentations to our dataset (translation, shifting, shearing, rotation), and we restrict the maximum depth for the budding tree to be 5.

% We evaluate the performance of our deep budding forest model on the MNIST dataset. We train the model layer by layer. Specifically, for layer L, and for each budding tree within the layer, we correspondingly construct a dataset with the same size as the original set by sampling with replacement. The motivation of this setting is the enable each budding tree sees a diversified version of the data, and their final concatenated representation will be more informative for further processing.  We also enhance the training set with different data augmentations (translation, shifting, shearing, rotating). We train a three layer deep model using budding tree/distributed budding tree as building blocks, each layer consists of 15 trees, for each budding tree we restrict its maximum depth to be 5.

Table \ref{table:comp} shows our results and comparison with results of other deep models. We can see that using distributed budding tree as building blocks for deep model is superior to the budding trees. For the feed forward forest, since it already has 2 layers, each layer with 2000 hard decision trees (with around 4096000 parameters in total), the performance still not as good as ours, which indicate that the deep budding forest can maintain compact model while guarantee performance. There is still more than one percent gap between our model and the Deep Conv Net, we think the reason is that the model doesn’t have effective feature extractor like convolutional filters, adding this function to our model will be our future work.

To show that the budding tree is able to learn more compact structures, Figure \ref{fig:fig3} displays the learned structure of a budding tree with depth 5, it can be seen that the tree is far from a complete tree, meaning that our growing tree is capable of learning compact tree topology that best suits the problem.

\section{Conclusion and Future Work}
In this paper, we extend the deep forest model by using a different building block - soft budding tree model.  Equipped with soft routing and leafness learning, the budding tree is able to learn a globally optimal structure by gradient descent method, thus is superior to hard decision trees in expressive power, we demonstrate this argument by showing that the deep budding forest model achieves comparable results on MNIST dataset to the feed forward deep forest model, but with much less trees at each layer, and less parameters overall.

There are several directions that we can explore to enhance the budding tree model. First, to incorporate the strength of convolution feature extraction into the model, so a richer span of representations can be extracted. Second,  there are better ways to form intermediate representations of each layer, rather than simple concatenation of the outputs, for example, by using dynamic routing like \cite{sabour2017dynamic}, the output of a budding tree in one layer can be sent to the next layer weighted by learnable attention weights.





%\bibliographystyle{IEEEtran}

\bibliographystyle{unsrt}
\bibliography{ICDM}

\begin{thebibliography}{10}

\bibitem{zhou2017deep}
Zhi-Hua Zhou and Ji~Feng.
\newblock Deep forest: Towards an alternative to deep neural networks.
\newblock {\em arXiv preprint arXiv:1702.08835}, 2017.

\bibitem{hettinger2017forward}
Chris Hettinger, Tanner Christensen, Ben Ehlert, Jeffrey Humpherys, Tyler
  Jarvis, and Sean Wade.
\newblock Forward thinking: Building and training neural networks one layer at
  a time.
\newblock {\em arXiv preprint arXiv:1706.02480}, 2017.

\bibitem{miller2017forward}
Kevin Miller, Chris Hettinger, Jeffrey Humpherys, Tyler Jarvis, and David
  Kartchner.
\newblock Forward thinking: Building deep random forests.
\newblock {\em arXiv preprint arXiv:1705.07366}, 2017.

\bibitem{wu2018deep}
Junru Wu, Yue Wang, Zhenyu Wu, Zhangyang Wang, Ashok Veeraraghavan, and Yingyan
  Lin.
\newblock Deep $ k $-means: Re-training and parameter sharing with harder
  cluster assignments for compressing deep convolutions.
\newblock {\em arXiv preprint arXiv:1806.09228}, 2018.

\bibitem{irsoy2014budding}
Ozan Irsoy, Olcay~Taner Yildiz, and Ethem Alpaydin.
\newblock Budding trees.
\newblock In {\em Pattern Recognition (ICPR), 2014 22nd International
  Conference on}, pages 3582--3587. IEEE, 2014.

\bibitem{irsoy2012soft}
Ozan Irsoy, Olcay~Taner Y{\i}ld{\i}z, and Ethem Alpayd{\i}n.
\newblock Soft decision trees.
\newblock In {\em Pattern Recognition (ICPR), 2012 21st International
  Conference on}, pages 1819--1822. IEEE, 2012.

\bibitem{irsoy2014distributed}
Ozan Irsoy and Ethem Alpayd{\i}n.
\newblock Distributed decision trees.
\newblock {\em arXiv preprint arXiv:1412.6388}, 2014.

\bibitem{huang2017densely}
Gao Huang, Zhuang Liu, Laurens Van Der~Maaten, and Kilian~Q Weinberger.
\newblock Densely connected convolutional networks.
\newblock In {\em CVPR}, volume~1, page~3, 2017.

\bibitem{ciresan2011flexible}
Dan~C Ciresan, Ueli Meier, Jonathan Masci, Luca Maria~Gambardella, and
  J{\"u}rgen Schmidhuber.
\newblock Flexible, high performance convolutional neural networks for image
  classification.
\newblock In {\em IJCAI Proceedings-International Joint Conference on
  Artificial Intelligence}, volume~22, page 1237. Barcelona, Spain, 2011.

\bibitem{sabour2017dynamic}
Sara Sabour, Nicholas Frosst, and Geoffrey~E Hinton.
\newblock Dynamic routing between capsules.
\newblock In {\em Advances in Neural Information Processing Systems}, pages
  3856--3866, 2017.

\end{thebibliography}

\end{document}